\UseRawInputEncoding
\documentclass[runningheads]{llncs}
\pdfoutput=1%For ARXIV
\usepackage{cite}

\usepackage[T1]{fontenc}
\usepackage{graphicx}
\usepackage{hyperref}
\usepackage{amsfonts}
\usepackage{amsopn}
\begin{document}
\title{Dynamic Structural Brain Network Construction by Hierarchical Prototype Embedding GCN using T1-MRI}

%\titlerunning{Abbreviated paper title}
% If the paper title is too long for the running head, you can set
% an abbreviated paper title here
%
% \author{Anonymous1\inst{1} \and
% Anonymous2\inst{2} \and
% Anonymous3\inst{3} \and
% % Anonymous4\inst{2} \and
% Anonymous5\inst{1}}
\author{Yilin Leng\inst{1,2} \and
Wenju Cui\inst{2,3} \and
Chen Bai\inst{2,3} \and
% Zirui chen\inst{4} \and
Zheng Yanyan\inst{4} \and
Jian Zheng\inst{2,3}}
\authorrunning{Y. Leng et al.}

% First names are abbreviated in the running head.
% If there are more than two authors, 'et al.' is used.
%
% \institute{Anonymous Organization1 \\
% % % \email{***@***.***} \and
% % Anonymous Organization2 \and
% Anonymous Organization3}
\institute{Institute of Biomedical Engineering, School of Communication and Information Engineering, Shanghai University, Shanghai, China \and the Medical Imaging Department, Suzhou Institute of Biomedical Engineering and Technology, Chinese Academy of Sciences, Suzhou, China \and School of Biomedical Engineering(Suzhou), Division of Life Sciences and Medicine, University of Science and Technology of China, Hefei, China \and Wenzhou Medical University, Wenzhou, China}

\maketitle              % typeset the header of the contribution
\begin{abstract}
Constructing structural brain networks using T1-weighted magnetic resonance imaging (T1-MRI) presents a significant challenge due to the lack of direct regional connectivity information.
Current methods with T1-MRI rely on predefined regions or isolated pretrained location modules to obtain atrophic regions, which neglects individual specificity. 
Besides, existing methods capture global structural context only on the whole-image-level, which weaken correlation between regions and the hierarchical distribution nature of brain connectivity.
We hereby propose a novel dynamic structural brain network construction method based on T1-MRI, which can dynamically localize critical regions and constrain the hierarchical distribution among them for constructing dynamic structural brain network. 
Specifically, we first cluster spatially-correlated channel and generate several critical brain regions as prototypes. 
Further, we introduce a contrastive loss function to constrain the prototypes distribution, which embed the hierarchical brain semantic structure into the latent space.
Self-attention and GCN are then used to dynamically construct hierarchical correlations of critical regions for brain network and explore the correlation, respectively. 
Our method is evaluated on ADNI-1 and ADNI-2 databases for mild cognitive impairment (MCI) conversion prediction, and acheive the state-of-the-art (SOTA) performance. 
Our source code is available at http://github.com/*******.
\end{abstract}
\section{Introduction}
T1-weighted magnetic resonance imaging (T1-MRI) is one of the indispensable medical imaging methods for noninvasive diagnosing neurological disorder \cite{frisoni2010clinical}. 
Existing approaches \cite{liu2018landmark, pan2021disease} based on T1-MRI focus on extracting region of interests (ROIs) to analyze structural atrophy information associated with disease progression. 
However, some works \cite{liu2018landmark, shao2020hypergraph, cui2022bmnet} heavily rely on manual defined and selected ROIs, which have limitations in explaining the individual brain specificity. 
To address this issue, Lian et al. \cite{lian2018hierarchical} localize discriminative regions by a pretrained module, where region localization and following feature learning cannot reinforce each other, resulting a coarse feature representation.
Additionally, as inter-regional correlations are unavailable in T1-MRI directly, most related works \cite{chen2022alzheimer, lian2020attention} ignore inter-regional correlations or replace them with a generalized global information. 
These conventional modular approaches have limitations in explaining high-dimensional brain structure information \cite{bullmore2009complex, van2011rich}.

Brain network is a vital method to analysis brain disease, which has been widely used in functional magnetic resonance imaging (fMRI) and diffusion tensor imaging (DTI). 
However, the structural brain network with T1-MRI is still underexplored due to the lack of direct regional connectivity.
Recent advances \cite{kipf2016semi, song2021graph, song2022multi, duran2022dual} in graph convolution neural networks (GCNs) have optimized brain networks construction with fMRI and DTI.
Given the successful application of GCN in these modalities, we think that it also has potential for construction of structural brain network using T1-MRI. 
% \deleted[id=1]{This will make it possible to extract correlations of structural atrophy information between brain regions.} % \cite{bullmore2009complex, van2011rich}
Current approaches \cite{chen2022adversarial, lei2020self, li2022joint, duran2022dual} to brain network construction involve the selection of ROIs and modeling inter-regional correlations, in which anatomical ROIs are employed as nodes, and inter-node correlations are modeled as edges. 
Some researches\cite{zhou2006hierarchical, meunier2009hierarchical} have demonstrated that brain connectivity displays hierarchical structure distribution, yet most GCN-based methods\cite{ye2020attention, li2022joint} treat all nodes equally and ignore the hierarchical nature of brain connectivity.
These structural brain networks are fixed and redundant, which may lead to coarse feature representation and suboptimal performance in downstream tasks. 

To address these issues, we propose novel \textbf{d}ynamic structural brain network construction method named \textbf{h}ierarchical \textbf{pro}totypes embedding \textbf{GCN} (DH-ProGCN) to dynamically construct disease-related structural brain network based on T1-MRI. 
Firstly, a prototypes learning method is used to cluster spatially-correlated channel and generate several critical brain regions as prototypes. 
Then, we introduce a contrastive loss function to constrain the hierarchical distribution among prototypes to obtain the hierarchical brain semantic structure embdedding in the latent space.
% for constructing dynamic structural brain network
% Then by hierarchical prototype learning, DH-ProGCN learns the dynamic representations that reflect hierarchical semantic structures in the latent space. 
After that, DH-ProGCN utilizes a self-attention mechanism to dynamically construct hierarchical correlations of critical regions for constructing structural brain network. GCN is applied  to explore the correlation of the structural brain network for Mild Cognitive Impairment (MCI) conversion prediction. 
% The key contributions of our approach are as follows: 
% 1) We propose a novel framework DH-ProGCN for dynamic structural brain network construction with T1-MRI; 
% 2) We introduce a hierarchical prototype learning method for dynamic critical brain regions learning and hierarchical semantic structure embedding of the brain in the latent space; 
We verify the effectiveness of DH-ProGCN on the Alzheimer’s Disease Neuroimaging Initiative-1 (ADNI-1) and ADNI-2 dataset.
DH-ProGCN achieves state-of-the-art (SOTA) performance for the the classification of progressive mild cognitive impairment (pMCI) and stable mild cognitive impairment (sMCI) based on T1-MRI.

\begin{figure}[ht]
\includegraphics[width=\textwidth]{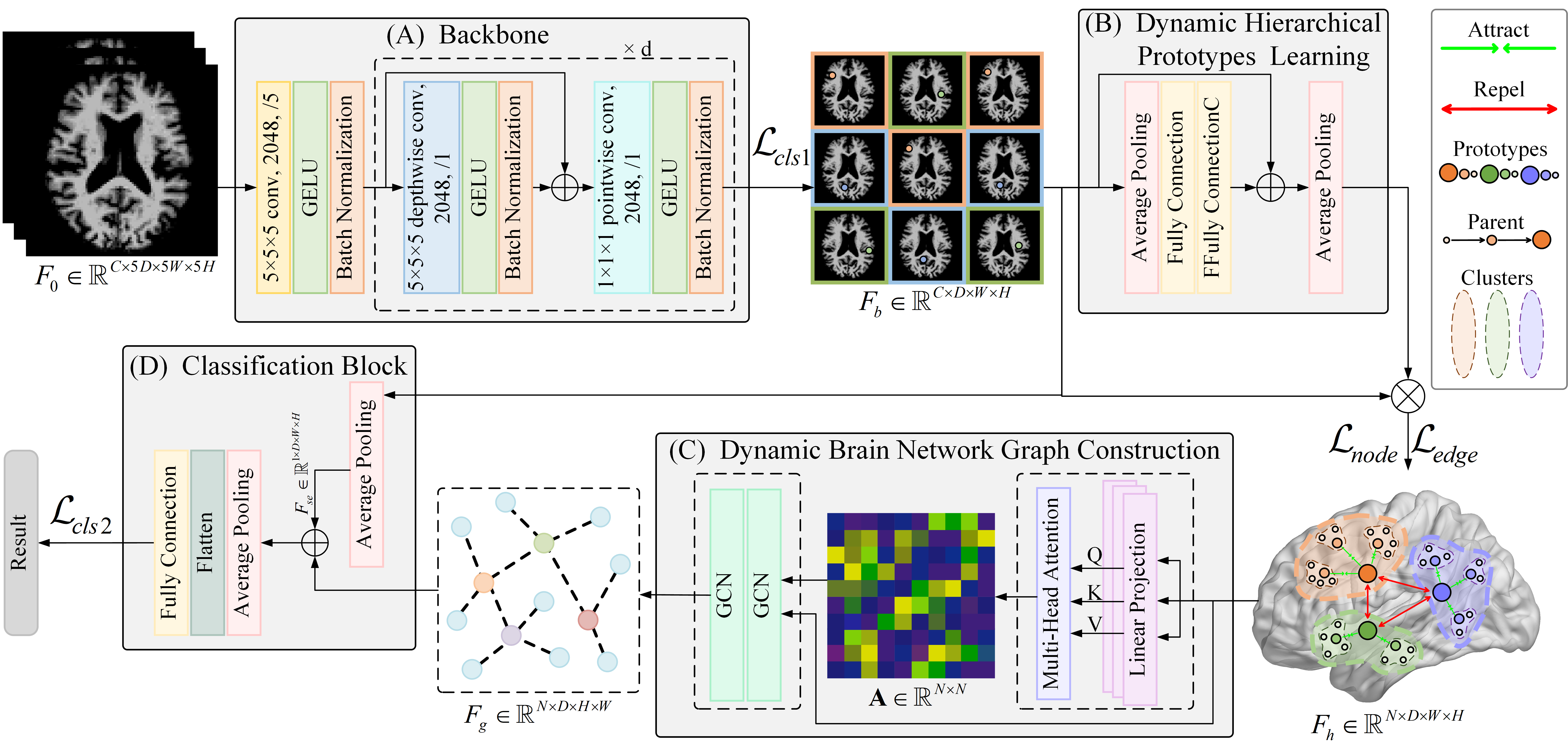}
\caption{The overall framework of the DH-ProGCN. (A) We first extract the feature $F_b$ via backbone, and assume that the featuremap of each channel represents different discriminative regions which are showed as images with different colors in $F_b$. (B) The hierarchical feature $F_h$ are then obtained by hierarchical clustering on the channel dimension. (C) We utilize a self-attention mechanism to model feature correlations matrix $A$ and learn the feature graph $F_g$ by a GCN. (D) $F_g$ and the global representation $F_b$ are concatenated for MCI conversion prediction.}\label{fig1}
\end{figure}

\section{Methods}

\subsection{Backbone}
% \subsubsection{Backbone.}
In this study, we utilize a Convmixer-like \cite{trockman2022patches} block as the backbone to achieve primary discriminative brain regions localization, which could provide a large enough channel dimension for subsequent channel clustering with relatively low complexity. 
Specifically, depicted in Fig.~\ref{fig1}(A), the backbone consists of a patch embedding layer followed by several full-convolution blocks. Patch embedding comprises a $5\times5\times5$ convolution, and the full convolution block comprikuaijses a $5\times5\times5$ depthwise convolution (grouped convolution with groups equal to the number of channels) and a pointwise convolution (kernel size is $1\times1\times1$) with 2048 channels. By the backbone, features of discriminative regions are finally extracted as $F_b\in {\mathbb{R}}^{C\times D\times H\times W}$, where $D$, $H$, $W$ and $C$ indicate depth, height, width and the number of channels, respectively.

\subsection{Dynamic Hierarchical Prototype Learning}
 
\subsubsection{Prototypes Definition.}
In this study, we regard feature maps of each channel as corresponding to the response of distinct brain regions relevant to tasks. Following \cite{zheng2017learning}, we utilize the location of each peak response as the channel information. Intuitively, a position vector composed of peak response coordinates of each channel is defined as the candidate prototype. Position vectors of training images can be obtained as following:
\begin{equation}
[t_{x}^1, t_{y}^1, t_{z}^1, t_{x}^2, t_{y}^2, t_{z}^2,\ldots, t_{x}^{\Omega}, t_{y}^{\Omega}, t_{z}^{\Omega}]
\end{equation}
where $[t_{x}^{i}$, $t_{y}^{i}$, $t_{z}^{i}]$ represents the peak response coordinate of the $i$-$th$ image and $\Omega$ represents the number of images in  the training set. $K$-means \cite{lloyd1982least} is used to achieve prototypes initialization.
Specifically, vectors of all channels are clustered to obtain $N$ sets of clusters $K = {\{k_{n}\}}_{n=1}^{N}$, and prototypes are defined as clustering centers $\Gamma = \{\gamma_{n}\}_{n=1}^{N}$ which are taken as $N$ critical regions for the discriminative localization (i.e., ROIs). $F_h\in \mathbb{R}^{N \times D \times H \times W}$ represents features of clustering centers.

\subsubsection{Dynamic Hierarchical Prototype Exploring.}

Inter-regional spatial connectivity is fixed, but the correlation between them is dynamic with disease progression. 
We argue that there are structural correlations between different regions, just as the complex hierarchical functional connectome in rich-clubs \cite{van2011rich} organization with fMRI. 
We therefore explore the hierarchical semantic structure of critical brain regions by the  hierarchical prototype clustering method.

Specifically, we start by using the initial prototypes as the first hierarchy clustering prototypes, denoted as $\Gamma^0 = \{\gamma_n^0\}_{n=1}^{N_0}$. 
Then, $K$-means is applied iteratively to obtain parent prototypes of the lower-hierarchy prototypes $\Gamma^{i-1} = \{\gamma_n^{i-1}\}_{ n=1}^{N_{i-1}}$, denoted as $\Gamma^i = \{\gamma_n^i\}_{ n=1}^{N_i}$, where $i$ represents the $i$-$th$ hierarchy and $N_i$ represents the number of clusters at $i$-$th$ hierarchy, corresponding to the cluster $K^i = {\{k_n^i\}}_{n=1}^{N_i}$. 
In this paper, $i$ is set as 2. The number of prototypes in the first, second and third hierarchy is set as 16, 8 and 4, respectively.

To facilitate optimal clustering of the network during training, we use two fully convolutional layers with two contrastive learning loss functions $\mathcal{L}_{node}$ and $\mathcal{L}_{edge}$ to approximate the clustering process. 
With $\mathcal{L}_{node}$, each channel clustering is enforced to become more compact inside and have significant inter-class differences with other clusterings, enabling all prototypes to be well separated:
% $\mathcal{L}_{node}$ enforces \textbf{channel clustering compact with significant inter-class differences}, enabling all prototypes to be well separated according to following operation:
\begin{equation}
    \mathcal{L}_{node}=\displaystyle -\frac{1}{L} \sum_{l=1}^{L}\sum_{n=1}^{N_l} \displaystyle \sum_{u \in K_n^l} \log \frac{\exp \left(u \cdot \gamma_n^l/ \phi_n^l\right)}{\sum_{i\neq n}^{N_l} \exp \left(u \cdot \gamma_i^l / \phi_n^l\right)}
\end{equation}

\begin{equation}
    \phi_n^l=\frac{\sum_{u \in K_n^l}\left\|u-\gamma_n^l\right\|_2}{|K_n^l| \cdot \log (|K_n^l|+\alpha)}
\end{equation}
Where $L$ is the total number of layers, and $N_l$ is the number of clusters in the $l$-$th$ layer. 
$K_n^l$, $\gamma_n^l$, and $\phi_n^l$ denote the set of all elements, the cluster center (prototype), and the estimation of concentration of the $n$-$th$ cluster in the $l$-$th$ layer, respectively.
$\alpha$ is a smoothing parameter to prevent small clusters from having overly-large $\phi$.

The cluster concentration $\phi$ measures the closeness of elements in a cluster. A larger $\phi$ indicates more elements in the cluster or smaller total average distance between all elements and the cluster center.
Ultimately, $\mathcal{L}_{node}$ compels all elements $u$ in $K_n^l$ to be close to their cluster center $\gamma_n^l$ and away from other cluster center at the same level.

Similarly, $\mathcal{L}_{edge}$ aims to embed the hierarchical correlation between clustering prototypes, which can be expressed as:
\begin{equation}
    \mathcal{L}_{edge}=\displaystyle -\frac{1}{L}\sum_{l=1}^{L-1}  \displaystyle \sum_{n=1}^{N_l} \log \frac{\exp \left(\gamma_n^l \cdot Parent(\gamma_n^l) / \tau\right)}{\sum_{i\neq n}^{N^l} \exp \left(\gamma_n^l \cdot \gamma_i^l / \tau\right)}
\end{equation}
$Parent(\gamma_n^l)$ represents the parent prototype of the prototype $\gamma_n^l$, and $\tau$ is a temperature hyper-parameter. $\mathcal{L}_{edge}$ forces all prototypes $\gamma^l$ in the $l$-$th$ layer to be close to their parent prototype and away from other prototypes within the same level.

\subsection{Brain Network Graph Construction and Classification}
Through Section 2.2, critical brain regions are clustered in a hierarchical semantic latent space. We hereby employ the prototypes regions as nodes and correlations between them as edges to construct structural brain network graphs as shown in Fig.~\ref{fig1}(C). 

We first apply a self-attention mechanism \cite{vaswani2017attention} to compute inter-region correlations to generate edges of the brain network. 
Then, the features $F_h$ is input to three separate fully connected layers to obtain three vectors: query, key, and value, which are used to compute attention scores $\mathbf{A} \in \mathbb{R}^{N \times N}$ between each pair of prototypes, followed by being used to weight the value vector and obtain the output of the self-attention layer as following operation:
% To the end, we obtain the correlation information (edges of the brain network graph) $\mathbf{A} \in \mathbb{R}^{N \times N}$ among all critical regions as following operation:
\begin{equation}
    \mathbf{A}=Attention(Q,K,V)=softmax(\frac{QK^T}{\sqrt[]{d_k}})V
\end{equation}
where $Q \in \mathbb{R}^{N \times d_k}$, $K \in \mathbb{R}^{N \times d_k}$, $V \in \mathbb{R}^{N \times N}$ denote query, key, and value, respectively. $d_k$ represents the dimension of $Q$, $K$. $N$ represents the number of critical regions, which is set as 16 in this paper. 

We then employ GCN to capture the topological interaction in the brain network graph and update features of nodes by performing the following operation:
\begin{equation}
GCN(\mathbf{X})=\operatorname{ReLU}\left(\hat{\mathbf{D}}^{-1 / 2} \hat{\mathbf{A}} \hat{\mathbf{D}}^{-1 / 2} \mathbf{X} \Theta\right)
\end{equation}
where $\hat{\mathbf{A}}=\mathbf{A}+\mathbf{I}$ is the adjacency matrix with inserted self-loops and $\mathbf{I}$ denotes an identity matrix. $\hat{\mathbf{D}}_{ii}=\sum_{j=0}\hat{\mathbf{A}}_{ij}$ is the diagonal degree matrix, and $\Theta$ represents learned weights. To prevent the network overfitting, we just use two GCN layers as the encoder to obtain the final graph feature $F_{g}\in \mathbb{R}^{N\times D \times H \times W}$.

To this end, the information of critical brain regions are fully learned. Notably, as prototypes are dynamic, constructed brain network graphs are also dynamic, rather than predefined and fixed. 
This allows DH-ProGCN to model and explore the individual hierarchical information, providing a more personalise brain network representation for every subject.

To achieve the classification, we perform channel squeezing on the backbone feature $F_b$ to obtain global features $F_{se} \in \mathbb{R}^{1 \times D \times H \times W}$, concatenate it with $F_g$ and input them into the classification layer.

\section{Experiments}

\subsection{Dataset}
The data we used are from two public databases: ADNI-1 (http://www.adni-info.org) \cite{petersen2010alzheimer}, and ADNI-2.
The demographic information of the subjects and preprocessing steps are shown in the supplemental material. The preprocessed images are finally resized to $91 \times 109 \times 91$ voxels. Through the quality checking, 305 images are left from ADNI-1 (197 for sMCI, 108 for pMCI), and 250 images are left from ADNI-2 (251 for sMCI, 99 for pMCI). Note that some subjects have two images or more in different times, and we only keep the earliest one. Following \cite{lian2018hierarchical}, we train DH-ProGCN on ADNI-1 and perform independent testing on ADNI-2.

\subsection{Implementation Details}
We first train backbone with 2048 channels in all layers to extract the output features $F_b$ with the cross-entropy loss $\mathcal{L}_{cls1}$. The cross-entropy loss $\mathcal{L}_{cls2}$ is used for the final classification. The overall loss function is defined as:
\begin{equation}
\mathcal{L}=\mathcal{L}_{cls1}+\mathcal{L}_{cls2}+\mathcal{L}_{node}+\mathcal{L}_{edge}
\end{equation}
where $\mathcal{L}_{node}$ and $\mathcal{L}_{edge}$ are explained in Section 2.2. Smooth parameter $\alpha = 10$ and temperature parameter $\tau = 0.2$ following \cite{chen2020improved}.

All blocks are trained by SGD optimizer with a momentum of 0.9 and weight decay of 0.001. The model is trained for 300 epochs with an initial learning rate of 0.01 that is decreased by a factor of 10 every 100 epochs. Five metrics, namely accuracy (ACC), sensitivity (SEN), specificity (SPE), and area under the curve (AUC), are used to evaluate the performance of the proposed model. We use Python based on the PyTorch package and run the network on a single NVIDIA GeForce 3090 24 GB GPU.

\section{Results}

\subsection{Comparing with SOTA Methods}
Six SOTA methods are used for comparison: 1) LDMIL \cite{liu2018landmark} captured both local information conveyed by patches and global information; 2) H-FCN \cite{lian2018hierarchical} implemented three levels of networks to obtain multi-scale feature representations which are fused for the construction of hierarchical classifiers; 3) HybNet \cite{lian2020attention} assigned the subject-level label to patches for local feature learning by iterative network pruning; 4) AD$^2$A \cite{guan2021multi} located discriminative disease-related regions by an attention modules; 5) DSNet \cite{pan2021disease} provided disease-image specificity to an image synthesis network; 6) MSA3D \cite{chen2022alzheimer} implemented a slice-level attention and a 3D CNN to capture subject-level structural changes.

Results in Table~\ref{tab2} show the superiority of DH-ProGCN over SOTA approaches for MCI conversion prediction. 
Specifically, DH-ProGCN achieves ACC of 0.849 and AUC of 0.845 tested on ADNI-2 by models trained on ADNI-1. 
It is worth noting that our method: 
1) needs no predefined manual landmarks, but achieves better diagnostic results than existing deep-learning-based MCI diagnosis methods; 
2) needs no pretrain network parameters from other tasks like AD diagnosis; 
3) introduces hierarchical distribution structure to connect regions and form region-based specificity brain structure networks, rather than generalizing the correlations between regions with global information.

% Table generated by Excel2LaTeX from sheet 'sota'
\begin{table}
  \centering
  \caption{Comparsion of our method with current SOTA methods for MCI conversion prediction on ADNI-2 obtained by the models trained on ADNI-1.}\label{tab2}
    \begin{tabular}{p{7em}p{3em}p{3em}p{3em}p{3em}}
    \hline
    Method & ACC & SEN & SPE & AUC \\
    \hline
    LDMIL  & 0.769  & 0.421  & 0.824  & 0.776  \\
    H-FCN & 0.809  & 0.526  & 0.854  & 0.781  \\
    HybNet & 0.827  & 0.579  & 0.866  & 0.793  \\
    AD$^2$A & 0.780  & 0.534  & 0.866  & 0.788  \\
    DSNet   & 0.762  & \textbf{0.770}  & 0.742  & 0.818  \\
    MSA3D  & 0.801  & 0.520  & 0.856  & 0.789  \\
    \textbf{DH-ProGCN}  & \textbf{0.849}  & 0.647  & \textbf{0.928}  & \textbf{0.845}  \\
    \hline
    \end{tabular}%
  \label{tab:addlabel}%
\end{table}%

\begin{figure}[ht]
\includegraphics[width=\textwidth]{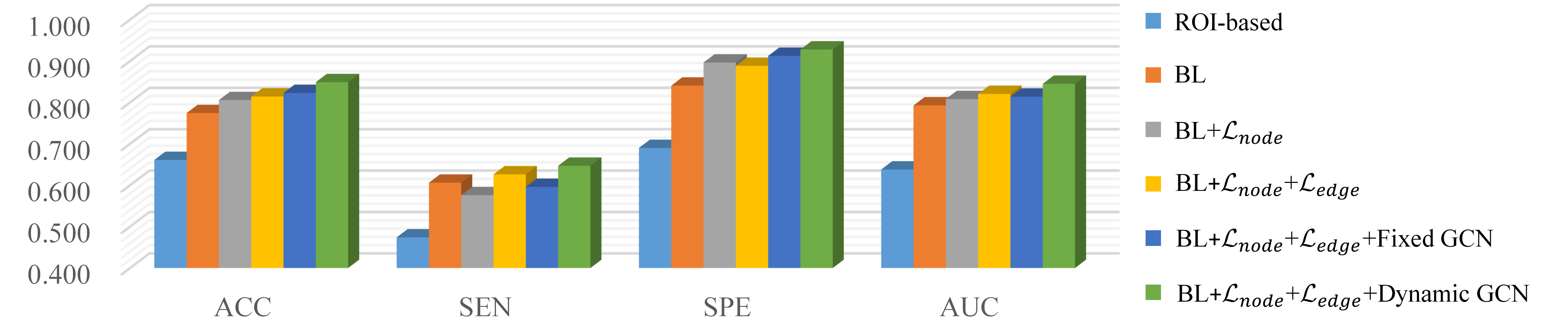}
\caption{Effects of each component of DH-ProGCN for MCI conversion prediction on ADNI-2 obtained by models trained on ADNI-1.}\label{fig2}
\end{figure}

\subsection{Ablation Study}

\subsubsection{Effect of dynamic prototype learning.}
To verify the effect of dynamic prototype clustering, we compare 1) ROI-based approach \cite{zhang2011multimodal}, 2) backbone without channel clustering (BL), 3) backbone with dynamic prototypes clustering (BL+$\mathcal{L}_{node}$). 
As shown in Fig.~\ref{fig2}, results indicate that dynamic prototype clustering outperforms the ROI-based and backbone on MCI conversion, and could generate better feature distributions for downstream brain images analysis tasks. 

\subsubsection{Effect of Hierarchical prototype learning.}
To evaluate the impact of hierarchical prototype learning, we compare backbone with flattened prototypes clustering (BL+$\mathcal{L}_{node}$), and hierarchical clustering (BL+$\mathcal{L}_{node}$+$\mathcal{L}_{edge}$). The results are presented in Fig.~\ref{fig2}. 
With the constraint strengthened on the distribution of regions, the results are progressively improved. This implies that it makes sense to introduce hierarchical semantics into the construction of structure brain networks.

\begin{figure}[ht]
\includegraphics[width=0.99\textwidth]{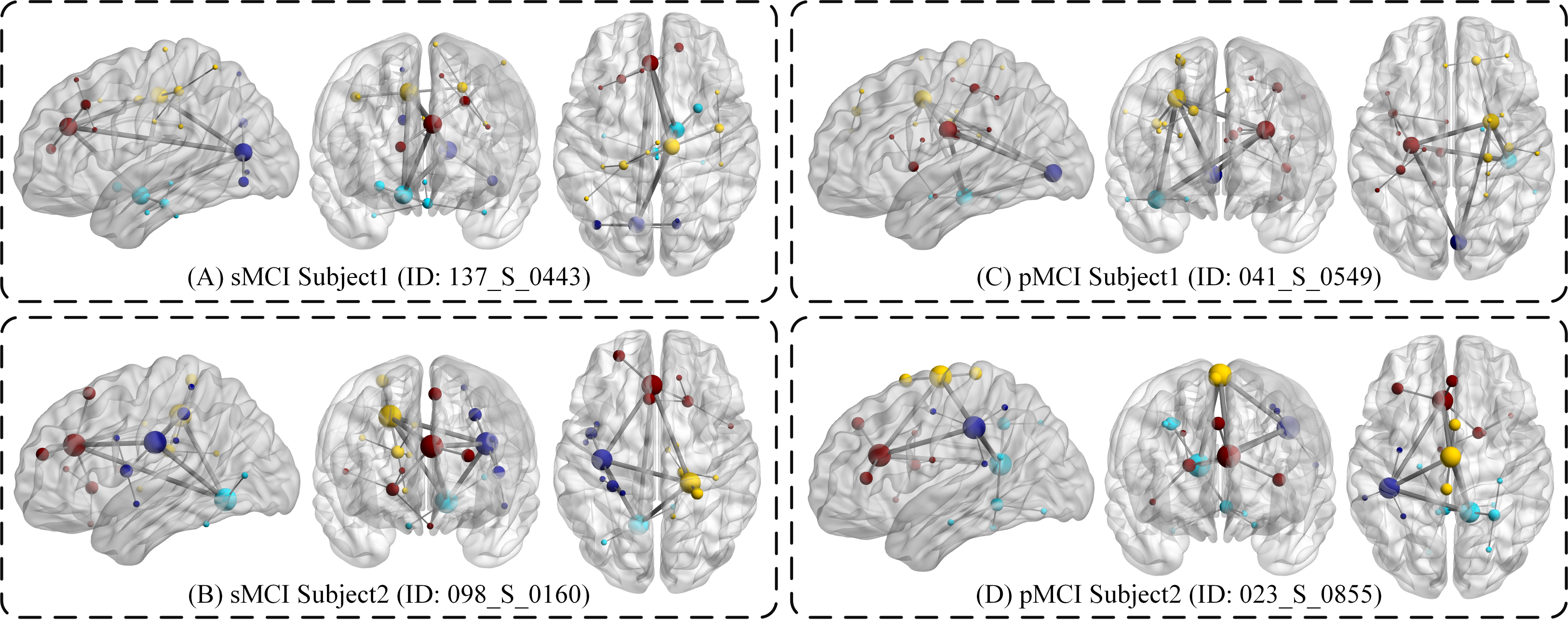}
\caption{Sagittal, coronal and axial views of connectome in hierarchical critical regions. (A)(B) represent brain network visualization of sMCI and (C)(D) represent pMCI subjects. Nodes correspond to critical regions i.e. prototypes, and edges are form the connectivity weight between nodes. The size of node increases with its hierarchy, and nodes with same color are clustered into the same parent prototype. Lower-hierarchy prototypes within cluster are closer to its parent prototypes, and higher-hierarchy prototypes between different clusters are closer than lower-hierarchy prototypes.}\label{fig3}
\end{figure}

\subsubsection{Effect of Dynamic Brain Network Construction.}
To verify whether our constructed dynamic brain network capability outperforms the fixed architecture, we obtained the fixed brain network graph by directly connecting all critical regions after obtaining hierarchical features and feeding them into the GCN for classification. The results are shown in Fig.~\ref{fig2}, where the dynamic brain network structure performs better, suggesting that the correlation between regions needs to be measured dynamically to construct a better brain network.

In addition, we visualize the sagittal, coronal and axial views of hierarchical critical regions and their connectome in Fig.~\ref{fig3}.
The left and right sub-figures represent brain network visualization of two sMCI and two pMCI subjects, respectively. 
% \added[id=1]{We co-locate the critical regions of the  }
% For different subjects, the critical regions we localize are different,
In general, critical regions and correlations are varied for different subjects, which means that our method is feasible for constructing individual brain networks according to the individuals specificity. 
Localized regions are roughly distributed in anatomically defined parahippocampal gyrus, superior frontal gyrus, and cingulate gyrus for different sMCI subjects, lingual gyrus right, and superior longitudinal fasciculus for different pMCI subjects, which agree with previous studies. \cite{frisoni2010clinical, chincarini2011local, dickerson2001mri}.

\section{Conclusion}
In this paper, we propose a novel dynamic structural brain network construction method named DH-ProGCN. 
DH-ProGCN could dynamically cluster critical brain regions by the prototype learning, implicitly encode the hierarchical semantic structure of the brain into the latent space by hierarchical prototypes embedding, dynamically construct brain networks by self-attention and extract topology features in the brain network by GCN. 
Experimental results show that DH-ProGCN outperforms SOTA methods on the MCI conversion task. 
Essentially, DH-ProGCN has the potential to model hierarchical topological structures in other kinds of medical images. 
In our future work, we will apply this framework to other kinds of modalities and neurological disorders.

\subsubsection{Acknowledgements} This work is partially supported by ********* and ********.

% Data used in preparation of this article was obtained from the Alzheimers Disease Neuroimaging Initiative (ADNI) database (http://www.adni-info.org/). The investigators within the ADNI contributed to the design and implementation of ADNI and/or provided data, but did not participate in analysis or writing of this report. A complete listing of ADNI investigators can be found at http://adni.loni.usc.edu/wp-content/uploads/how\_to\_apply/ADNI.

%
% ---- Bibliography ----
%
% BibTeX users should specify bibliography style 'splncs04'.
% References will then be sorted and formatted in the correct style.
%
% \bibliographystyle{splncs04}
\bibliographystyle{unsrt}
\bibliography{main.bbl}

\end{document}